\documentclass[10pt,twocolumn,letterpaper]{article}

\PassOptionsToPackage{table}{xcolor}
\usepackage[pagenumbers]{cvpr} %

\usepackage[T1]{fontenc}
\usepackage{graphicx}
\usepackage{booktabs}
\usepackage{tabularx}
\usepackage{float}
\usepackage[normalem]{ulem}
\definecolor{tabfirst}{HTML}{FF7F7F}
\definecolor{tabsecond}{HTML}{FFBF7F}
\definecolor{tabthird}{HTML}{FFFF7F}

\newcommand{\stock}[1]{\textcolor{black!40}{#1}}

\usepackage{tikz}
\newcommand{\qualtrim}{0.037}
\newlength{\quallab}\newlength{\qualw}\newlength{\qualh}
\newcommand{\qualsetup}[1][3]{%
  \setlength{\quallab}{0.065\columnwidth}%
  \setlength{\qualw}{\dimexpr(\columnwidth-\quallab-#1pt)/#1\relax}%
  \setlength{\qualh}{\dimexpr\qualw-\qualtrim\qualw-\qualtrim\qualw\relax}%
  \footnotesize\setlength{\tabcolsep}{0pt}\renewcommand{\arraystretch}{0}}
\newcommand{\qualstrut}{\rule[-.1ex]{0pt}{1.65ex}}
\newcommand{\qualrow}[1]{\makebox[\quallab]{\rotatebox[origin=c]{90}{#1}}}
\newcommand{\qualinlayside}{l}
\newcommand{\qualcrop}[3][]{\raisebox{-.5\height}{\includegraphics[width=\qualw,clip,
  trim=0 {\the\dimexpr\qualtrim\dimexpr#3bp\relax\relax} 0 {\the\dimexpr\qualtrim\dimexpr#3bp\relax\relax}]{#2}%
  \if\relax\detokenize{#1}\relax\else\llap{\makebox[\qualw][\qualinlayside]{\setlength{\fboxsep}{1.5pt}%
    \raisebox{\depth}{\colorbox{black}{\textcolor{white}{\scriptsize\strut#1}}}}}\fi}}

\newcommand{\qualview}[6][.92]{\begin{tikzpicture}
  \path[use as bounding box] (0,0) rectangle (#1\qualw,#1\qualw);
  \node[anchor=south west,inner sep=0] (v) {\includegraphics[width=#1\qualw]{#2}};
  \begin{scope}[x={(v.south east)},y={(v.north west)}]
    \draw[black,line width=2.05pt] (#3,#4) rectangle (#5,#6);
    \draw[white,line width=0.94pt] (#3,#4) rectangle (#5,#6);
  \end{scope}\end{tikzpicture}}

\tikzset{cell label/.style={fill=black, text=white, inner xsep=5pt, inner ysep=3pt, text depth=0pt, font=\scriptsize}}

\newcommand{\labelimg}[3][]{\begin{tikzpicture}[inner sep=0pt, outer sep=0pt]
  \node[anchor=south west] (img) at (0,0) {\includegraphics[width=#2]{#3}};
  \if\relax\detokenize{#1}\relax\else\node[cell label, anchor=south east] at (img.south east) {#1};\fi
\end{tikzpicture}}

\newlength{\gridw}

\definecolor{cvprblue}{rgb}{0.21,0.49,0.74}
\usepackage[pagebackref,breaklinks,colorlinks,allcolors=cvprblue]{hyperref}

\title{Fast 3D Gaussian Ray Tracing for Wide-Angle Reconstruction}

\author{Miguel Lemus\textsuperscript{1\dag} \qquad
Yohan Poirier-Ginter\textsuperscript{1,2\dag} \qquad
Jean-Fran\c{c}ois Lalonde\textsuperscript{1}\\[2pt]
\textsuperscript{1}Universit\'e Laval \qquad
\textsuperscript{2}Centre Inria d'Universit\'e C\^ote d'Azur \qquad
\textsuperscript{\dag}Equal contribution
}

\makeatletter
\patchcmd{\@maketitle}{\vskip .375in}{\vskip 6pt}{}{\PackageError{preamble}{title patch 1 failed}{}}
\patchcmd{\@maketitle}{\vspace*{24pt}}{\vspace*{9.1pt}}{}{\PackageError{preamble}{title patch 2 failed}{}}
\makeatother

\begin{document}

\twocolumn[{%
\renewcommand\twocolumn[1][]{#1}%
\maketitle
\begin{center}
    \vspace{-1.2em}
    \includegraphics[width=\textwidth]{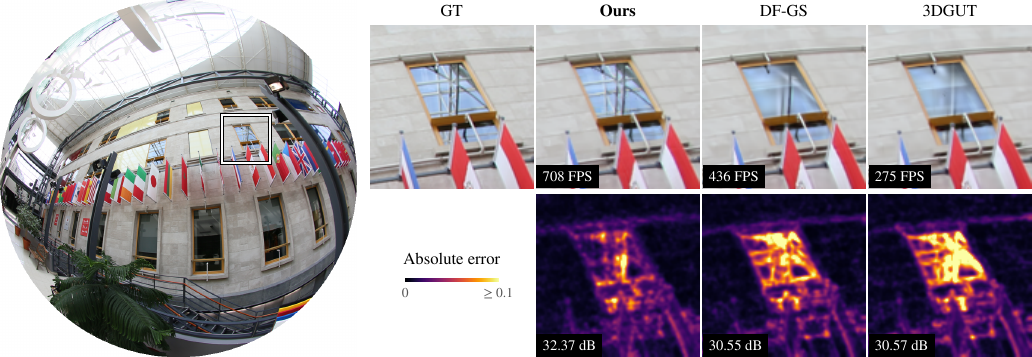}
    \captionof{figure}{
    Our generic Gaussian ray tracer outperforms specialized wide-angle rasterizers while supporting arbitrary lens distortion models. 
    Left: a held-out fisheye photograph of \emph{atrium}, a scene from our new circular fisheye benchmark \emph{Fish\&Chips}. Right: the framed region rendered by each method (top) and its absolute error (bottom), alongside frame rates and PSNR averaged over all Fish\&Chips scenes.}
    \label{fig:teaser}
\end{center}%
}]

\begin{abstract}
3D Gaussian Splatting's rasterization enables fast, high-quality novel-view synthesis, but is poorly suited to wide-angle lenses. While Gaussian ray tracing natively supports arbitrary camera models, early methods were slower, leading prior work to adapt rasterization instead. In this work, we present the first Gaussian ray tracer to match the performance of the original 3DGS method and, building on this advance, revisit wide-angle Gaussian reconstruction through ray tracing. Our generic method surpasses state-of-the-art rasterizers specialized for wide-angle fisheye benchmarks while supporting arbitrary lens distortion models, and outperforms omnidirectional methods by training directly on unstitched dual-fisheye inputs. We further use it to systematically evaluate more complex lens distortion functions, and find that accounting for non-central parallax can substantially improve reconstruction in scenes with near-field occluders. To demonstrate this, we introduce a new dataset of seven circular fisheye scenes. Our results suggest that ray tracing is not only a natural fit, but also a practical solution for wide-angle Gaussian reconstruction. All code and data will be released upon publication.
\end{abstract}

\vspace{0pt plus 1fill} %

\section{Introduction}
\label{sec:intro}

3D Gaussian Splatting (3DGS)~\cite{3dgs} is widely adopted for novel-view synthesis as its explicit Gaussian-based formulation enables high-quality radiance-field rendering in real time. This efficiency comes from its rasterization-based pipeline~\cite{zwicker}, which projects 3D Gaussians onto the image plane and composites them in screen space. Much of 3DGS's popularity comes from its speed, and much of that speed comes from rasterization.

Unfortunately, rasterization inherently assumes a pinhole camera model. Therefore, to handle distortion and lenses with larger fields of view, rasterization-based approaches must train on rectified input images~\cite{3dgs,gsplat,nerfstudio}. While this approach is effective for mild distortion, rectifying wide-angle images stretches them excessively~\cite{rectification-survey}.

3D Gaussians can also be rendered by ray tracing~\cite{3dgrt}, which natively supports arbitrary lens models. In practice, as early ray tracers were substantially slower than splatting, methods focused on adapting rasterization to wide-angle imagery. Unfortunately, the camera-projection linearization used by 3DGS becomes inaccurate in this setting~\cite{fisheye-gs,directfisheye-gs,360-gs,omnigs,odgs}; exact 3D rasterizers can resolve this issue but involve other compromises~\cite{3dgut,spags,3dgeer}.

In this work, we introduce the first Gaussian ray tracer to match the speed and quality of the original 3DGS method. In particular, we accelerate ray tracing by introducing a differentiable cost regularizer that explicitly accounts for rendering speed in the optimization objective. This allows for cost-aware optimization strategies that find better local performance trade-offs: specifically, we use it to guide both the pruning of Gaussians and the adjustment of per-Gaussian hyperparameters. Our state-of-the-art ray tracer renders an order of magnitude faster than 3DGRT~\cite{3dgrt} at its PSNR.

Building on this advance, we revisit wide-angle Gaussian rendering through ray tracing, and show that a generic method trained directly on distorted fisheye images can outperform Gaussian rasterizers explicitly designed for wide-angle reconstruction. Further, it can outperform omnidirectional methods by training directly on the dual-fisheye images captured by 360\textdegree{} cameras. As it does not compromise rendering speed, our method calls into question the preference for rasterization in wide-angle reconstruction.

Finally, we use our method's support for arbitrary lens models to systematically evaluate more complex distortion functions, including one that accounts for lens non-centrality, \ie, tracing rays from different origins. We find that accounting for non-central parallax can substantially improve reconstruction in scenes with near-field occluders, which we demonstrate in a new circular fisheye benchmark comprising seven scenes with near-field perspective effects.

Our contributions are:
\begin{itemize}
\item A state-of-the-art Gaussian ray tracer, the first to match 3DGS in both quality and speed through differentiable cost-aware regularization;
\item State-of-the-art wide-angle Gaussian reconstruction on standard fisheye and omnidirectional datasets with an unspecialized approach;
\item A systematic evaluation of lens distortion models, showing that modeling non-central effects can increase reconstruction quality in the presence of near-field occluders;
\item A benchmark of seven challenging circular fisheye scenes featuring occasional near-field occluders.
\end{itemize}

Our results show that ray tracing is not only a natural fit, but a practical solution for wide-angle Gaussian reconstruction and real-time rendering. All code and data will be released upon publication.

\section{Related work}
\label{sec:related}

We review Gaussian splatting and ray tracing (\cref{sec:related:3dgs}) and their adaptations to wide-angle lenses (\cref{sec:related:wide-angle}).

\subsection{Gaussian splatting and ray tracing}
\label{sec:related:3dgs}

\begin{figure}[t]
\centering
\includegraphics[width=\columnwidth]{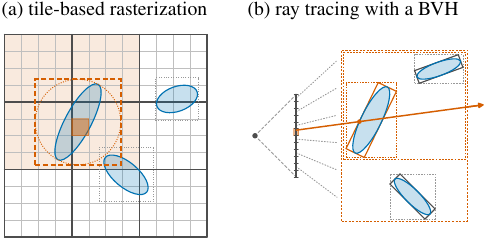}
\caption{Typical rasterizers coarsely assign Gaussians to tiles and process them collectively using approximations, while ray tracers evaluate each pixel exactly and independently.}
\label{fig:raster-vs-rt}
\end{figure}

At a high level, Gaussian splatting~\cite{3dgs} and ray tracing~\cite{3dgrt} both render anisotropic 3D Gaussians through three main steps: assigning Gaussians to the pixels they influence (attribution), ordering them front-to-back (sorting), and alpha-blending their evaluated contributions (compositing). 

\paragraph{Tile-based rasterization.} 3DGS's rasterizer partitions the image into fixed-size tiles. Since perspective projection does not preserve Gaussianity, it flattens each primitive onto the image plane using a local affine approximation. The resulting 2D footprints are assigned to tiles using conservative screen-space bounds and sorted by centroid depth (\cref{fig:raster-vs-rt}a); each pixel blends its tile's Gaussians in this shared order.

This efficient approach unfortunately requires undistorted input images. Moreover, the affine approximation introduces errors for Gaussians spanning large image regions, and sharing the depth ordering causes popping as overlapping Gaussians swap order during camera motion; both issues worsen at wider fields of view. 

\paragraph{Hardware-accelerated ray tracing.} 3DGRT~\cite{3dgrt} and other Gaussian ray tracers~\cite{raygauss,raygaussx,ever,raysplats,redisplats,stochastic-ray-1,stochastic-ray-2,epbrr,grtx} use a hardware-accelerated~\cite{optix} bounding volume hierarchy (BVH) to identify the Gaussians intersected by each ray (\cref{fig:raster-vs-rt}b). These are sorted per pixel and evaluated exactly in 3D during compositing, avoiding rasterization's approximations and supporting arbitrary lens models. 

While per-pixel processing can be slower than splatting, GRay~\cite{gray} approaches the speed of 3DGS by reducing ray-Gaussian intersections via dense initialization~\cite{edgs} and scale regularization that results in smaller Gaussians with limited volumetric overlap. This narrows the performance gap with Gaussian splatting, which we close in this work.

\paragraph{Improved rasterization.} StopThePop~\cite{stopthepop} resolves popping through per-pixel sorting but remains somewhat expensive. To reduce affine projection error, Huang et al.~\cite{optimal-gaussian-splatting} project Gaussians onto view-aligned tangent planes. Addressing both limitations, Hahlbohm et al.'s perspective-accurate rasterizer~\cite{perspective-accurate} evaluates Gaussians exactly in 3D and limits per-pixel sorting to the nearest few Gaussians, blending the rest approximately. These methods approach ray tracing's accuracy more efficiently for perspective images, but remain relatively difficult to extend to wide-angle lenses.

\subsection{Wide-angle adaptations}
\label{sec:related:wide-angle}

Several methods already adapt 3D Gaussians to wide-angle images, but nearly all opt for rasterization. \Cref{tab:method-comparison} summarizes how wide-angle rasterizers either fall short of 3DGS's speed or sacrifice ray tracing's capabilities.

\begin{table}[H]
\centering
\footnotesize
\setlength{\tabcolsep}{1.5pt}
\begin{tabularx}{\columnwidth}{@{}>{\raggedright\arraybackslash}Xcccccc@{}}
\toprule
 & 3DGUT & DF-GS & 3DGEER & SPaGS & UTS & Ours \\
\midrule
\textbf{Exact tile attribution} & $\times$ & $\checkmark$ & $\checkmark$ & $\checkmark$ & $\checkmark$ & $\checkmark$ \\
\textbf{Exact 3D evaluation} & $\checkmark$ & $\times$ & $\checkmark $ & $\checkmark$ & $\times$ & $\checkmark$ \\
\textbf{Accurate pinhole render} & $\checkmark$ & $\times$ & $\checkmark$ & $\checkmark$ & $\checkmark$ & $\checkmark$ \\
\textbf{Supports any distortion} & $\checkmark$ & $\checkmark$ & $\checkmark$ & $\times$ & $\checkmark$ & $\checkmark$ \\
\textbf{Resampling-free} & $\checkmark$ & $\checkmark$ & $\times/\checkmark$ & $\checkmark$ & $\times$ & $\checkmark$ \\
\textbf{Popping-free} & $\times/\checkmark$ & $\times$ & $\times$ & $\checkmark$ & $\times$ & $\checkmark$ \\
\textbf{3DGS-level speed} & $\checkmark/\times$ & $\checkmark$ & $\times$ & $\checkmark$ & $\times$ & $\checkmark$ \\

\bottomrule
\end{tabularx}
\caption{Capabilities of wide-angle Gaussian rendering methods. Rasterization-based methods make trade-offs to maintain 3DGS-level speed; we present a new ray tracer that avoids such compromises without sacrificing speed. }
\label{tab:method-comparison}
\end{table}

\begin{figure}[t]
\centering
\includegraphics[width=\columnwidth]{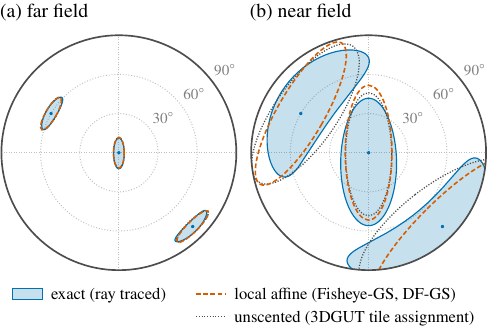}
\caption{Image-space footprint of the same 3D Gaussian in an equidistant fisheye image at incident angles of 0\textdegree{}, 60\textdegree{}, and 80\textdegree{}, (a) far from and (b) near the camera. The approximations used by many rasterizers drift towards the rim where the exact footprint bends with the projection.}
\label{fig:footprint}
\end{figure}

\paragraph{Fisheye methods.} Fisheye-GS~\cite{fisheye-gs} extends 3DGS to fisheye cameras by replacing the perspective projection and its affine approximation with their equidistant fisheye counterparts. DirectFisheye-GS~\cite{directfisheye-gs} (DF-GS) additionally incorporates radial coefficients from the Kannala--Brandt~\cite{kannala-brandt} distortion model. These methods support various lens models without undistortion, but linearizing the projection can introduce errors under strong deformation (\cref{fig:footprint}) or when switching to a pinhole camera after training~\cite{unitrisplat}.

\paragraph{Omnidirectional methods.} Omnidirectional approaches take equirectangular 360\textdegree{} panoramas as input~\cite{rectification-survey}, typically stitched from dual-fisheye images~\cite{dual-fisheye-stiching}. Approximate methods include OmniGS~\cite{omnigs}, 360-GS~\cite{360-gs}, and ODGS~\cite{odgs}, which likewise implement the affine approximation in equirectangular space. The state-of-the-art SPaGS~\cite{spags} extends perspective-accurate rendering~\cite{perspective-accurate} to equirectangular cameras by evaluating Gaussians exactly in 3D. However, it requires specialized footprint bounds and fisheye stitching, which introduces parallax artifacts that persist in the reconstruction~\cite{seam360gs}; our method avoids these by training directly on dual-fisheye images.

\paragraph{General-purpose methods.} 3DGUT~\cite{3dgut} evaluates Gaussians exactly in 3D but assigns them to tiles using coarse bounds from projected sigma points, assuming limited distortion (\cref{fig:footprint}). It exhibits popping unless it uses slower per-pixel sorting~\cite{stopthepop}. UniTriSplat~\cite{unitrisplat} (UTS) renders to a uniform HEALPix sphere~\cite{healpix}, with resampling trading off aliasing against rendering cost. 3DGEER~\cite{3dgeer} makes both evaluation and bounding exact, evaluating each Gaussian in closed form along the pixel ray and bounding it with a frustum derived from its 3D covariance. Its default equiangular grid requires resampling to and from the camera image; we use its released implementation's direct Kannala--Brandt rendering~\cite{kannala-brandt} in our comparisons, which significantly outperforms the paper's formulation.

\paragraph{Ray tracing methods.} Unlike the preceding methods, ray tracing supports arbitrary wide-angle camera models without modifying the renderer beyond camera-ray generation. FullCircle~\cite{fullcircle} recently applied 3DGRT directly to raw dual-fisheye captures, demonstrating its practicality for stitchless omnidirectional reconstruction, but retained its relatively low frame rates. Other Gaussian ray tracers~\cite{3dgrt,raygaussx,ever,raysplats,redisplats,gray} likewise do not match 3DGS's quality and speed. In this work, we present the first Gaussian ray tracer to do so and to outperform rasterization approaches in several wide-angle reconstruction tasks.

\paragraph{Other works.}
NeRF-based methods~\cite{nerf,instant-ngp,mip-nerf-360} naturally accommodate arbitrary camera models through ray marching, motivating numerous wide-angle approaches~\cite{perf,omni-nerf,360fusionnerf,omninerf,omnivoxel,msi-nerf,fbinerf}; these achieve high quality, but fall short of real-time rendering. Wide-angle reconstruction has also been explored in self-calibrating~\cite{self-cali-gs,sc-omnigs} and pose-free approaches~\cite{pfgs360}. We defer further discussion of these approaches to the supplemental.

\begin{figure*}[t]
\centering
\includegraphics[width=\textwidth]{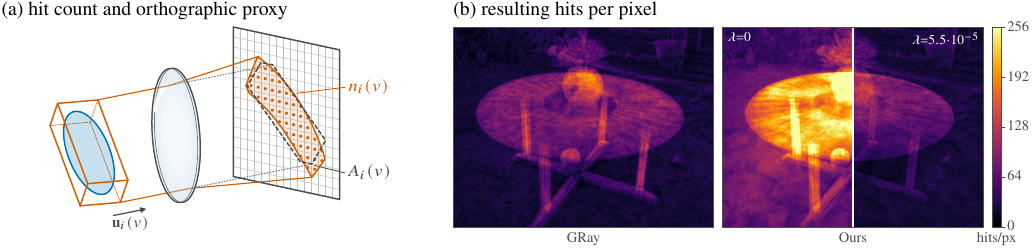}
\caption{Hit-count regularization. (a) The pixels of view $v$ whose rays hit the bounding box of Gaussian $i$ (dots) give its realized cost $n_i(v)$; as the box's footprint bends under lens distortion, we instead differentiate the area $A_i(v)$ of the box's orthographic projection along the direction $\mathbf{u}_i(v)$ from the camera to the Gaussian center (dashed). (b) Bounding-box hits per pixel on a Mip-NeRF 360 \emph{garden} test view for GRay and for our method, shown both with ($\lambda{=}5.5{\cdot}10^{-5}$) and without ($\lambda{=}0$) hit-count regularization. }
\label{fig:hits}
\end{figure*}

\section{Method}
\label{sec:method}

We introduce the first Gaussian ray tracer to match the performance of 3DGS (\cref{sec:method:matching-splatting}) and use it to model fisheye lenses natively (\cref{sec:method:fisheye}), further enabling reconstruction beyond central projection (\cref{sec:method:beyond}). 

\subsection{Cost-aware Gaussian ray tracing}
Our key idea is to make reconstruction aware of the BVH traversal costs through a differentiable proxy objective. This regularization allows optimization, pruning, and hyperparameter tuning to jointly balance computational overhead against quality.

\paragraph{Hit-count regularization.} Inspired by the surface area heuristic~\cite{pbrt}, which estimates BVH traversal cost from bounding-box area averaged over uniform directions, we seek to minimize the mean projected area of the Gaussians' oriented bounding boxes over the training views.

As computing bounding-box footprints under lens distortion is non-trivial, we instead use the realized hit count $n_i(v)$ of Gaussian $i$ in view $v$ to measure cost. This count is non-differentiable, so we approximate its relative change by that of the orthographic projected area $A_i(v)$ (\cref{fig:hits}):
\begin{equation}
\frac{\Delta n_i}{n_i}\approx\frac{\Delta A_i}{A_i}
\quad\Longrightarrow\quad
\nabla n_i \approx n_i\nabla\log A_i\,.
\end{equation}
This approximation yields the hit-count regularizer
\begin{equation}
\mathcal{L}_{\mathrm{hits}}
=
\mathbb{E}_v\!\smash[t]{\left[\frac{1}{N_{\mathrm{rays}}(v)}\sum_i n_i(v)\log A_i(v)\right]}\vphantom{\frac{1}{N_{\mathrm{rays}}}},
\end{equation}
where $N_{\mathrm{rays}}(v)$ is the number of rays sampled in view $v$ and $n_i(v)$ is held constant during differentiation. We add this term to the standard photometric objective:
\begin{equation}
\mathcal{L}
=
\mathcal{L}_{\mathrm{photo}}
+
\lambda\mathcal{L}_{\mathrm{hits}}\,,
\end{equation}
where $\lambda$ balances quality and rendering speed.

\begin{figure*}[t]
\centering
\includegraphics[width=\textwidth]{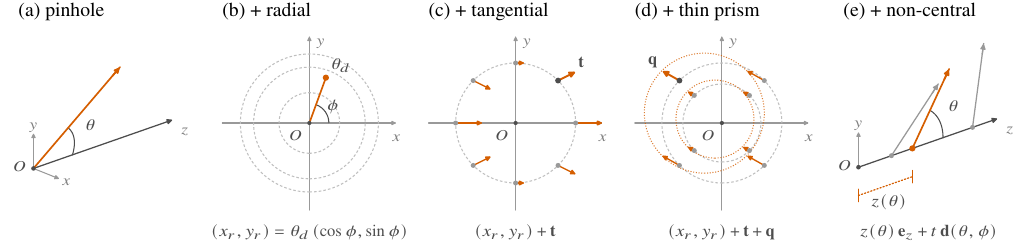}
\caption{Components of our fisheye distortion model, added one at a time. (a) In a pinhole camera, all rays leave the center $O$, at an incident angle $\theta$ from the optical axis $z$. (b) Radial distortion maps $\theta$ to the image radius $\theta_d$ and preserves the azimuth $\phi$. (c) Tangential distortion $\mathbf{t}$ varies with both radius and azimuth. (d) Thin-prism distortion $\mathbf{q}$ shifts all points at equal radius by the same offset. (e) Non-centrality moves each ray's origin along the optical axis by $z(\theta)$.}
\label{fig:lens}
\end{figure*}

\paragraph{Leave-one-out pruning.} We improve pruning by comparing what a Gaussian contributes with what it costs to render. We first measure contribution by the image change that removing a Gaussian causes along a ray:
\begin{equation}
  \Delta_i = \big\| w_i \mathbf{c}_i - \tfrac{\alpha_i}{1-\alpha_i}\mathbf{D}_i \big\|_1,
\end{equation}
where $w_i$, $\mathbf{c}_i$, and $\alpha_i$ are its blending weight, color, and opacity, and $\mathbf{D}_i$ is the color composited behind it. Our pruning score then balances $\Delta_i(v)$, the mean of $\Delta_i$ over the rays of view $v$, against hits per ray:
\begin{equation}
  s_i = \mathbb{E}_v\!\left[\Delta_i(v) - w_{\mathrm{prune}}\,\frac{n_i(v)}{N_{\mathrm{rays}}(v)}\right].
\end{equation}
Finally, we define the pruning weight $w_{\mathrm{prune}}$ as a function of $\lambda$ so that a single parameter controls the overall trade-off; further details are in the supplemental.

\paragraph{Differentiable hyperparameter tuning.} Our hit-count regularization provides a balancing force that lets us adjust hyperparameters automatically and locally: we replace GRay's fixed global quality hyperparameters with learnable per-Gaussian values. GRay controls Gaussian support with an exponent $n_{\mathrm{expo}}$ and an alpha threshold $\alpha_{\min}$, fixed at $2$ and $0.01$, respectively. To these we add a per-Gaussian \emph{squareness} parameter $p$ (initially $2$) by replacing the Euclidean norm with $\|\mathbf{x}\|_p=\left(\sum_i |x_i|^p\right)^{1/p}$; larger values of $p$ expand the support towards the corners of the bounding box without increasing traversal cost. We initialize these to their global values and optimize them jointly with the reconstruction.

\paragraph{Improved implementation.}
We improve GRay with:
\begin{itemize}
    \item a faster backward pass that realigns the per-pixel hit lists across
    the threads of a warp, further merging gradients with intrinsics to amortize atomic writes;
    \item improved dense initialization leveraging RoMa~v2~\cite{romav2} with multi-view triangulation and hole filling, which also replaces the brittle world-space point
    cloud fusion step with a screen-space approach invariant to scene scale;
    \item a low-resolution warm-up stage that benefits from ray tracing's higher speeds at low resolution.
\end{itemize}
The supplemental details these improvements alongside other minor adjustments.

\label{sec:method:matching-splatting}

\subsection{Native fisheye reconstruction}
\label{sec:method:fisheye}
Ray tracing natively supports expressive camera models and allows us to train directly on distorted images and evaluate the impact of modeling various distortion effects. Specifically, we experiment with the Kannala--Brandt~\cite{kannala-brandt} model extended with tangential~\cite{tangential} and thin-prism~\cite{thin-prism} distortion, as well as non-centrality~\cite{non-central} (\cref{fig:lens}). Unlike rasterization, our ray-tracing approach handles these models exactly without any change to the rendering pipeline.

\paragraph{Radial distortion.} The radial component (\cref{fig:lens}b) maps the incident ray angle $\theta$ to the distorted radius $\theta_d$ as
\begin{equation}
\theta_d
=
\theta
\left(
1 + \sum_{i=0}^{K} k_i\,\theta^{2(i+1)}
\right)\,.
\end{equation}
The radial coordinates become $(x_r, y_r) = \theta_d\,(\cos\phi, \sin\phi)$, where $\phi$ is the azimuthal angle of the incident ray.

\paragraph{Asymmetric distortion.} Tangential distortion (\cref{fig:lens}c) models lens decentering using coefficients $p_0$ and $p_1$,
\begin{equation}
\mathbf{t} = 2\,(p_0x + p_1y)\,(x, y) + \theta^2\,(p_0, p_1)\,,
\end{equation}
where $(x, y) = \theta\,(\cos\phi, \sin\phi)$. Thin-prism distortion (\cref{fig:lens}d) adds the asymmetric displacement $\mathbf{q} = \theta^2\,(s_0, s_1)$.
The distorted coordinates are $(x_d, y_d) = (x_r, y_r) + \mathbf{t} + \mathbf{q}$.

\begin{figure}
\centering
{\scriptsize\setlength{\tabcolsep}{0pt}\renewcommand{\arraystretch}{0}%
\newcommand{\ncview}[1]{\begin{tikzpicture}[inner sep=0pt]
  \node[anchor=south west] (v) {\includegraphics[width=\ncw]{figures/parallax/view_#1.png}};
  \path[use as bounding box] (v.south west) rectangle (v.north east);
  \node[anchor=north west] (z) at ([shift={(1.03pt,-1.03pt)}]v.north west) {\includegraphics[width=.54\ncw]{figures/parallax/zoom_#1.png}};
  \begin{scope}[x={(v.south east)},y={(v.north west)}]
    \foreach \c/\w in {black/2.05pt,white/.94pt}{%
      \draw[\c,line width=\w] (.7831,.3968) rectangle (.9867,.6009);
      \draw[\c,line width=\w] (z.south west) rectangle (z.north east);}
  \end{scope}\end{tikzpicture}}%
\newlength{\ncw}\setlength{\ncw}{\dimexpr(\columnwidth-3pt)/4\relax}%
\begin{tabular*}{\columnwidth}{@{}c@{\extracolsep{\fill}}ccc@{}}
\rule[-2.2pt]{0pt}{7.6pt}Central & 8\,cm & 14\,cm & 20\,cm \\
\ncview{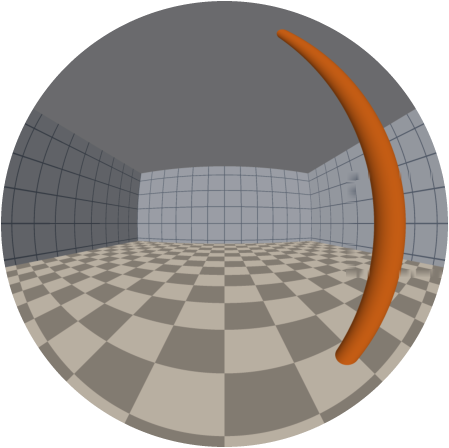} & \ncview{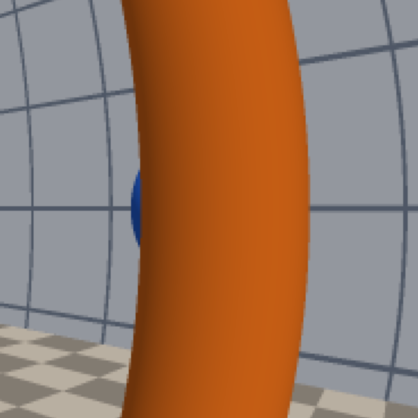} & \ncview{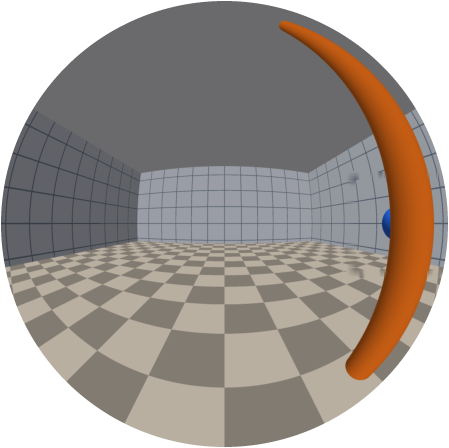} & \ncview{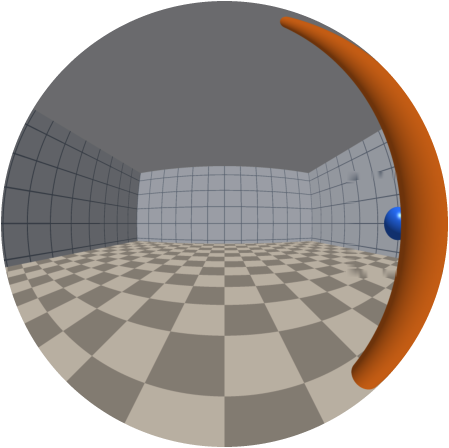}
\end{tabular*}}
\caption{Non-central parallax for exaggerated shifts $z(\theta_{\max})$.}
\label{fig:noncentral-parallax}
\end{figure}

\paragraph{Non-centrality.} Non-central models have been shown to model fisheye lenses more accurately, especially when scene elements are close to the camera~\cite{non-central}. These effects are best modeled via ray tracing, by varying the ray origin along the optical axis with the incident angle (\cref{fig:lens}e):
\begin{equation}
\mathbf{r}(t;\theta,\phi)
=
z(\theta)\,\mathbf{e}_z
+
t\,\mathbf{d}(\theta,\phi)\,,
\end{equation}
where $\mathbf{e}_z$ is the optical-axis unit vector and $\mathbf{d}(\theta,\phi)$ is the unit ray direction in camera coordinates. We define $z(\theta)=g_2\theta^2+g_4\theta^4$ similar to~\cite{non-central}. As the resulting parallax can reveal otherwise occluded objects (\cref{fig:noncentral-parallax}), it cannot be modeled with image undistortion. 

\subsection{Beyond a single center of projection}
\label{sec:method:beyond}
Unlike specialized rasterizers, ray tracing supports arbitrary camera models and complex imaging effects that extend beyond a single center of projection. For instance, omnidirectional reconstruction typically stitches dual-fisheye images into a panorama, which ray tracing can avoid.

\paragraph{Multi-camera imaging: unstitched panoramas.} Consumer 360\textdegree{} cameras typically stitch two opposing fisheye images into a single panorama~\cite{ricoh_theta_stitcher,insta360_stitching}. Treating it as a central image collapses the two lens centers into one, which can bake warping and stitching artifacts into the reconstruction~\cite{seam360gs}. Much like FullCircle~\cite{fullcircle}, we instead reconstruct directly from the original distorted fisheye images, preserving both lens centers and avoiding stitching. Our method reaches rendering speeds competitive with state-of-the-art omnidirectional rasterizers without sacrificing quality.

\paragraph{Depth of field and rolling shutter.} Ray tracing readily supports both depth of field and rolling shutter. The supplemental presents additional experiments demonstrating these effects, which our code release will support.

\begin{figure}[t]
\centering
{\scriptsize\setlength{\tabcolsep}{0pt}\renewcommand{\arraystretch}{0}%
\begin{tabular*}{\columnwidth}{@{}c@{\extracolsep{\fill}}c@{}}
\rule[-2.2pt]{0pt}{7.6pt}\textsc{Workshop} & \textsc{Library} \\
\includegraphics[width=.478\columnwidth]{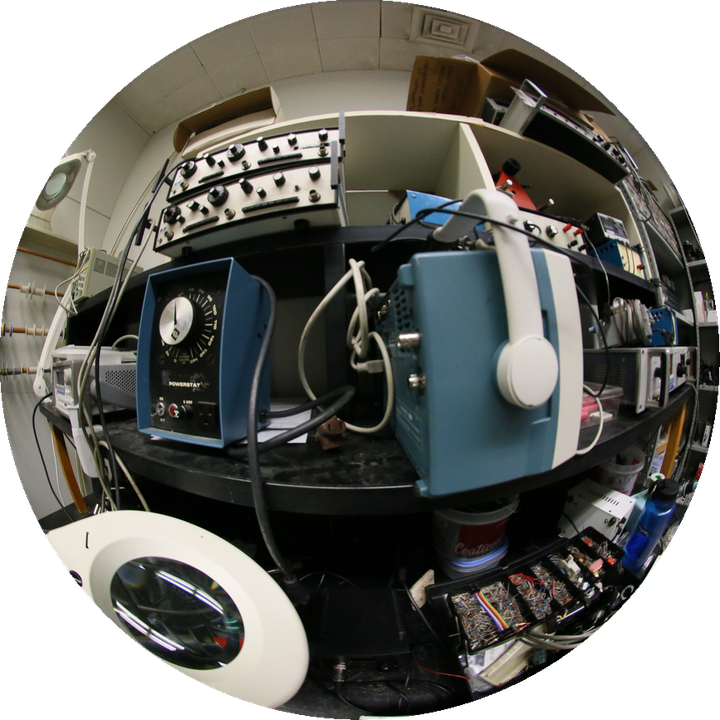} &
\includegraphics[width=.478\columnwidth]{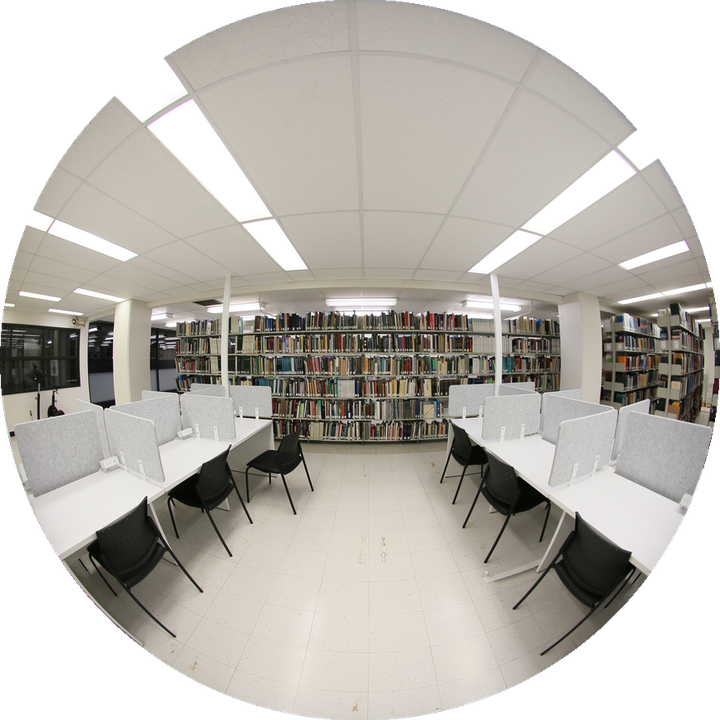}
\end{tabular*}}\\[3pt]
\newcommand{\ncbar}[3]{%
  \fill[cvprblue] (0,-#1-0.36) rectangle (#3,-#1+0.36);
  \node[right,inner sep=1pt,xshift=1.5pt,fill=white,font=\scriptsize] at (#3,-#1) {+#3};
  \node[left,inner sep=3pt,font=\scriptsize] at (0,-#1) {\textsc{#2}};}
\begin{tikzpicture}[x=.42\columnwidth,y=7.7pt]
  \draw[black!40,densely dashed,line width=.5pt] (0.78,-7.6) -- (0.78,-0.4);
  \node[right,inner sep=2pt,font=\scriptsize,text=black!40] at (0.78,-7) {mean +0.78};
  \ncbar{1}{Exhibition}{1.58}\ncbar{2}{Workshop}{1.48}\ncbar{3}{Reception}{0.69}\ncbar{4}{Classroom}{0.52}
  \ncbar{5}{Tunnel}{0.48}\ncbar{6}{Library}{0.47}\ncbar{7}{Atrium}{0.23}
  \draw[line width=.5pt] (0,-0.4) -- (0,-7.6);
  \node[inner sep=2pt,align=center,font=\scriptsize] at (1.33,-4.9) {PSNR gain (dB) from the\\non-central lens model};
\end{tikzpicture}
\caption{Top: samples from our Fish\&Chips scenes, which contain both near and far objects; near-field occluders cause non-central parallax. Bottom: PSNR gain from our non-central lens model on each Fish\&Chips scene.}
\label{fig:fishchips-nearfield}
\end{figure}

\section{Fisheye \& Challenging Perspectives dataset}
\label{sec:benchmark}

Existing multi-view fisheye datasets avoid close foreground objects, limiting non-central parallax by construction. To measure the importance of non-central lens models, we therefore introduce the Fish\&Chips (Fisheye \& Challenging Perspectives) dataset, a new benchmark of seven indoor scenes featuring occasional near-field occluders (\cref{fig:fishchips-nearfield}). We captured our scenes around a university campus using a Sigma 8\,mm f/3.5 EX DG circular fisheye lens, for which prior calibration reports an axial entrance-pupil shift of approximately 10\,mm at 90\textdegree{} off-axis~\cite{erwin-calib}.

\section{Experiments}

We first evaluate our Gaussian ray tracer in isolation on pinhole benchmarks (\cref{sec:experiments:raytracing}), followed by native wide-angle reconstruction on circular fisheye captures (\cref{sec:experiments:fisheye}) and unstitched dual-fisheye captures (\cref{sec:experiments:omni}). All timings were measured on an RTX 4090.

\subsection{State-of-the-art Gaussian ray tracing}
\label{sec:experiments:raytracing}

\begin{figure}[t]
\centering
\includegraphics[width=\columnwidth]{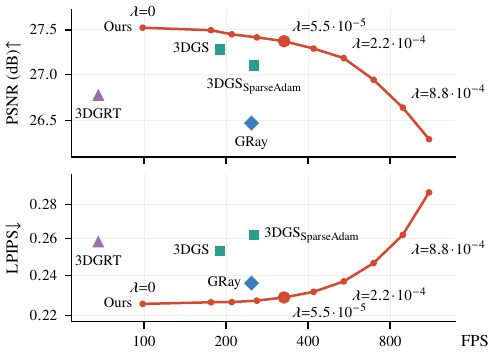}
\caption{Quality-speed Pareto fronts on the 13-scene 3DGS benchmark (pinhole projection model) when varying the regularization parameter $\lambda$.}
\label{fig:pareto}
\end{figure}

We compare our cost-aware Gaussian ray tracer with 3DGS~\cite{3dgs} on its 13-scene benchmark~\cite{mip-nerf-360,tanks-and-temples,deep-blending}; this pinhole comparison isolates the ray-tracing improvements of \cref{sec:method:matching-splatting}. \Cref{tab:pinhole-raytracing} shows our ray tracer matches the original 3DGS in quality, rendering speed, and optimization time (excluding initialization), while \cref{fig:pareto} shows the quality-speed trade-off of the regularization parameter $\lambda$; our method can reach nearly $12\times$ the rendering speed of 3DGRT at equal PSNR. Further combining our method with PPISP~\cite{ppisp}, Spherical Voronoi~\cite{spherical-vornoi}, and a weight EMA reaches 28.59\,dB PSNR at 345\,FPS.

\subsection{Distorted fisheye reconstruction}
\label{sec:experiments:fisheye}

We evaluate fisheye reconstruction on the FIORD~\cite{fiord}, Fisheye Zip-NeRF~\cite{zip-nerf}, and our new Fish\&Chips benchmarks. All methods are evaluated on held-out views using the same 180\textdegree{} field-of-view mask, and we compute PSNR, SSIM~\cite{ssim}, and LPIPS~\cite{lpips} directly on masked images.
    
\begin{table}[t!]
\centering
{\footnotesize
\setlength{\tabcolsep}{1pt}
\begin{tabularx}{\linewidth}{@{}>{\raggedright\arraybackslash}Xcccccc@{}}
\toprule
Method / variant & PSNR$_\uparrow$ & SSIM$_\uparrow$ & LPIPS$_\downarrow$ & FPS$_\uparrow$ & \#G$_\downarrow$ & Time$_\downarrow$ \\
\midrule
3DGS~\cite{3dgs} & 27.28 & 0.836 & 0.253 & 190 & 2.49M & 18:13 \\
3DGS\textsubscript{SparseAdam}~\cite{taming-3dgs} & 27.10 & 0.831 & 0.262 & 253 & 2.25M & 06:18 \\
3DGRT~\cite{3dgrt} & 26.77 & 0.828 & 0.258 & 68 & 3.24M & 55:01 \\
GRay~\cite{gray} & 26.47 & 0.819 & 0.236 & 248 & 1.52M & 05:40 \\
Ours ($\lambda{=}5.5{\cdot}10^{-5}$) & \textbf{27.37} & \textbf{0.845} & \textbf{0.229} & \textbf{327} & \textbf{0.83M} & \textbf{04:51} \\
\midrule
\enspace w/o Hit-count reg. & 27.52 & 0.848 & 0.226 & 99 & 0.91M & 07:05 \\
\enspace w/o Leave-one-out pruning & 27.38 & 0.845 & 0.218 & 261 & 2.39M & 06:31 \\
\enspace w/o Differentiable tuning & 26.98 & 0.838 & 0.244 & 350 & 0.76M & 03:57 \\
\bottomrule
\end{tabularx}\par}
\caption{Pinhole reconstruction performance on 3DGS's 13-scene benchmark and ray-tracer component ablations (\cref{sec:method:matching-splatting}). }
\label{tab:pinhole-raytracing}
\end{table}

\begin{table*}[t]
\centering
\resizebox{\textwidth}{!}{%
\begin{tabular}{@{}l*{18}{c}@{}}
\toprule
Method & \multicolumn{6}{c}{Fish\&Chips} & \multicolumn{6}{c}{FIORD~\cite{fiord}} & \multicolumn{6}{c}{Zip-NeRF~\cite{zip-nerf}} \\
\cmidrule(lr){2-7}\cmidrule(lr){8-13}\cmidrule(l){14-19}
 & PSNR$_\uparrow$ & SSIM$_\uparrow$ & LPIPS$_\downarrow$ & FPS$_\uparrow$ & \#G$_\downarrow$ & Time$_\downarrow$
 & PSNR$_\uparrow$ & SSIM$_\uparrow$ & LPIPS$_\downarrow$ & FPS$_\uparrow$ & \#G$_\downarrow$ & Time$_\downarrow$
 & PSNR$_\uparrow$ & SSIM$_\uparrow$ & LPIPS$_\downarrow$ & FPS$_\uparrow$ & \#G$_\downarrow$ & Time$_\downarrow$ \\
\midrule
\stock{DF-GS~\cite{directfisheye-gs} (stock)} & \stock{24.62} & \stock{0.895} & \stock{0.176} & \stock{688} & \stock{393k} & 
 & \stock{26.44} & \stock{0.853} & \stock{0.231} & \stock{630} & \stock{475k} & 
 & \stock{24.34} & \stock{0.827} & \stock{0.337} & \stock{757} & \stock{482k} &  \\
\stock{3DGUT~\cite{3dgut} (stock)} & \stock{26.69} & \stock{0.935} & \stock{0.141} & \stock{363} & \stock{103k} & 
 & \stock{26.89} & \stock{0.869} & \stock{0.220} & \stock{303} & \stock{216k} & 
 & \stock{24.08} & \stock{0.826} & \stock{0.342} & \stock{328} & \stock{496k} &  \\
\stock{3DGEER~\cite{3dgeer} (stock)} & \stock{27.31} & \stock{0.943} & \stock{0.120} & \stock{56} & \stock{2844k} & 
 & \stock{26.11} & \stock{0.861} & \stock{0.209} & \stock{60} & \stock{4099k} & 
 & \stock{24.53} & \stock{0.837} & \stock{0.319} & \stock{295} & \stock{604k} &  \\
DF-GS~\cite{directfisheye-gs} (improved) & 30.55 & 0.954 & 0.108 & 436 & 501k & 14:30
 & 31.69 & 0.935 & 0.136 & 285 & 969k & 17:48
 & 26.99 & 0.858 & 0.287 & 383 & 1030k & 15:06 \\
3DGUT~\cite{3dgut} (improved) & 30.57 & 0.954 & 0.112 & 275 & 460k & 08:24
 & 31.73 & 0.939 & 0.139 & 190 & 963k & 13:18
 & 26.66 & 0.858 & 0.295 & 246 & 1002k & 09:54 \\
3DGEER~\cite{3dgeer} (improved) & 31.80 & 0.960 & 0.104 & 138 & 394k & 47:24
 & 31.95 & 0.937 & 0.139 & 99 & 949k & 30:39
 & 26.99 & 0.863 & 0.279 & 188 & 1022k & 20:12 \\
Ours & \textbf{32.37} & \textbf{0.966} & \textbf{0.086} & \textbf{708} & \textbf{384k} & \textbf{03:12}
 & \textbf{32.47} & \textbf{0.940} & \textbf{0.113} & \textbf{657} & \textbf{739k} & \textbf{07:12}
 & \textbf{27.04} & \textbf{0.872} & \textbf{0.234} & \textbf{438} & \textbf{993k} & \textbf{08:18} \\
\midrule
\enspace w/o Tan. and thin prism & 32.37 & 0.966 & 0.086 & 704 & 386k & 
 & 32.43 & 0.940 & 0.113 & 658 & 740k & 
 & 26.97 & 0.871 & 0.235 & 436 & 995k &  \\
\enspace w/o Non-central & \underline{31.59} & \underline{0.960} & \underline{0.090} & 709 & 387k & 
 & 32.48 & 0.941 & 0.113 & 662 & 739k & 
 & 26.99 & 0.871 & 0.235 & 439 & 980k &  \\
\bottomrule
\end{tabular}}
\caption{Fisheye reconstruction comparisons showing our method reaches state-of-the-art quality and speed, alongside ablations for different lens distortion components. Overall, tangential and thin-prism distortion do not affect performance, but accurately modeling lens non-centrality is key in scenes with objects close to the camera (see the underlined ablation on our Fish\&Chips dataset). }
\label{tab:native-fisheye-results}
\end{table*}

\paragraph{Calibration.} We calibrate cameras with COLMAP~\cite{colmap}, modified to support non-central distortion. We first reconstruct with the radial-only \texttt{OPENCV\_FISHEYE} model (as a rig when applicable), then repeat global refinement with the extra distortion terms enabled (\texttt{THIN\_PRISM\_FISHEYE} and/or our non-central extension). Crucially, we also refine the principal point at this stage, which significantly improves quality and keeps the tangential and thin-prism terms from compensating for its offset. On FIORD, our calibration alone improves the unmodified 3DGEER and DF-GS baselines by 2.1 and 0.6\,dB PSNR, respectively, mostly from two scenes with inaccurate released intrinsics (in \emph{building\_out}, the released principal point differs from our estimate by 50 pixels).

\begin{figure}
\centering
{\renewcommand{\qualtrim}{0}\renewcommand{\qualinlayside}{r}\qualsetup[4]
\begin{tabular}{@{}c@{\hskip1pt}c@{\hskip1pt}c@{\hskip1pt}c@{\hskip1pt}c@{}}
\qualstrut & Test view & Ground truth & w/ Non-central & w/o Non-central \\\addlinespace[2pt]
\qualrow{\textsc{Exhibition}} & \raisebox{-.5\height}{\qualview[1]{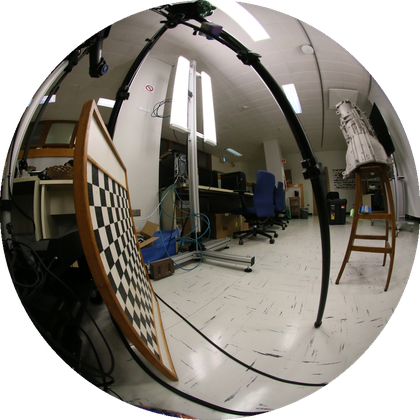}{0.1721}{0.3033}{0.3938}{0.5248}} & \qualcrop{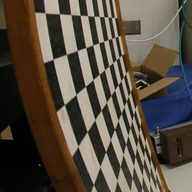}{192} & \qualcrop[+10.0 dB]{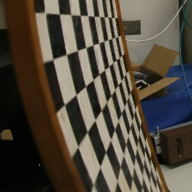}{192} & \qualcrop{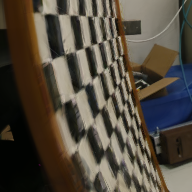}{192}\\\addlinespace[1pt]
\qualrow{\textsc{Workshop}} & \raisebox{-.5\height}{\qualview[1]{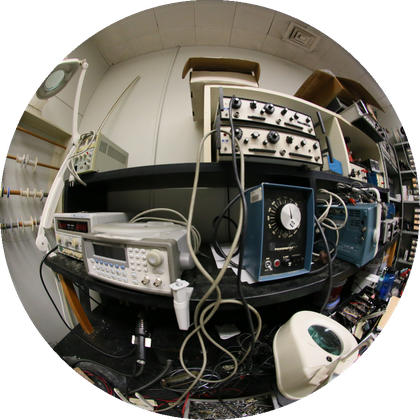}{0.5986}{0.3495}{0.7832}{0.5340}} & \qualcrop{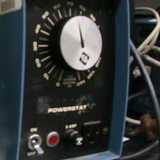}{160} & \qualcrop[+7.1 dB]{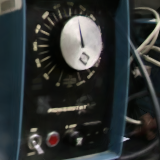}{160} & \qualcrop{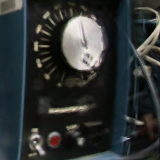}{160}\\\addlinespace[1pt]
\qualrow{\textsc{Reception}} & \raisebox{-.5\height}{\qualview[1]{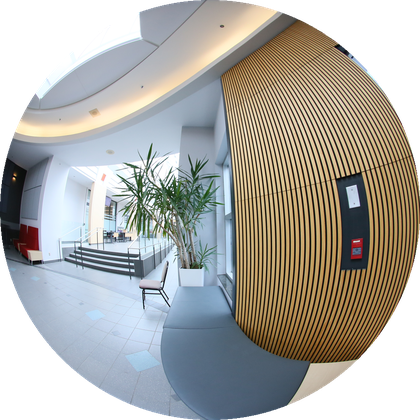}{0.4867}{0.1260}{0.7087}{0.3480}} & \qualcrop{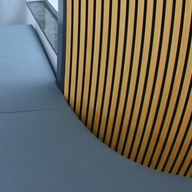}{192} & \qualcrop[+7.7 dB]{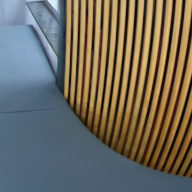}{192} & \qualcrop{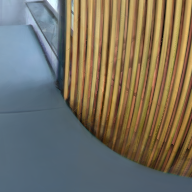}{192}
\end{tabular}}
\caption{Our method with and without non-central lens modeling on held-out Fish\&Chips views: without non-centrality, near-field objects degrade significantly. Insets give the PSNR gain on crops.}
\label{fig:noncentral-qualitative}
\end{figure}

\paragraph{Configuration.} We compare against 3DGUT~\cite{3dgut}, DF-GS~\cite{directfisheye-gs}, and 3DGEER~\cite{3dgeer} with their default code and calibrations at first, forming the (stock) rows in \cref{tab:native-fisheye-results}. The (improved) rows fix implementation issues to ensure consistent clipping at ultra-wide angles and use our improved calibrations. Finally, we adjust 3DGEER's densification to prevent uncontrolled growth, and lower densification thresholds per dataset so every baseline exceeds our Gaussian count. 

Our method on Fish\&Chips uses the same hyperparameters as the pinhole comparisons; on the larger FIORD and Zip-NeRF scenes, we disable warm-up, increase the batch size to 4, and double the number of dense initialization views; our method still trains faster than the baselines.

Following DF-GS~\cite{directfisheye-gs}, all methods include an exponential moving average of the Gaussian weights with decay $0.999$. We also account for vignetting by multiplying renders of all methods with a learnable radial vignette $V_c(r) = 1 + b_{1,c}r^2 + b_{2,c}r^4 + b_{3,c}r^6$, where $r$ is the distance to the principal point divided by the image mask radius and where the coefficients are shared across all views.
 
\begin{table}[t]
\centering
\footnotesize
\setlength{\tabcolsep}{2pt}
\begin{tabularx}{\linewidth}{@{}>{\raggedright\arraybackslash}Xcccccc@{}}
\toprule
Method & PSNR$_\uparrow$ & SSIM$_\uparrow$ & LPIPS$_\downarrow$ & FPS$_\uparrow$ & \#G$_\downarrow$ & Time$_\downarrow$ \\
\midrule
UniTriSplat~\cite{unitrisplat} & 25.45 & 0.820 & 0.380 & 31 & 672k & $\approx 1\mathrm{h}$ \\
SPaGS~\cite{spags} & 25.55 & 0.824 & 0.334 & 825 & 484k & 07:05 \\
FullCircle~\cite{fullcircle} & 28.53 & 0.892 & 0.268 & 282 & 489k & 14:43 \\
Ours & \textbf{28.80} & \textbf{0.895} & \textbf{0.262} & \textbf{1644} & \textbf{308k} & \textbf{02:34} \\
\bottomrule
\end{tabularx}
\caption{Omnidirectional reconstruction performance on the FullCircle dataset~\cite{fullcircle}.}
\label{tab:native-pano-results}
\end{table}

\begin{figure*}[!t]
\begin{minipage}[t]{\columnwidth}
\centering
{\qualsetup
\begin{tabular}[t]{@{}c@{\hskip1pt}c@{\hskip1pt}c@{\hskip1pt}c@{}}
\qualstrut & \textsc{Workshop} & \textsc{Classroom} & \textsc{Atrium} \\\addlinespace[2pt]
 & \qualview{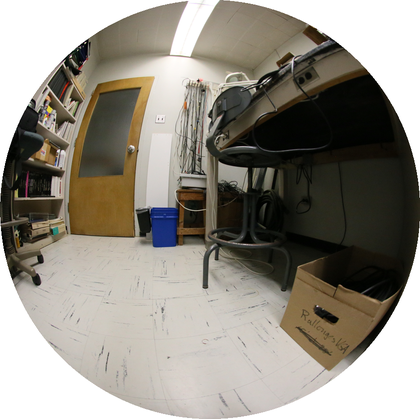}{0.6791}{0.1269}{0.8603}{0.3085} & \qualview{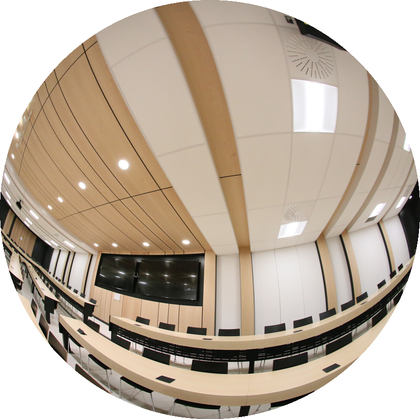}{0.2538}{0.0158}{0.4448}{0.2073} & \qualview{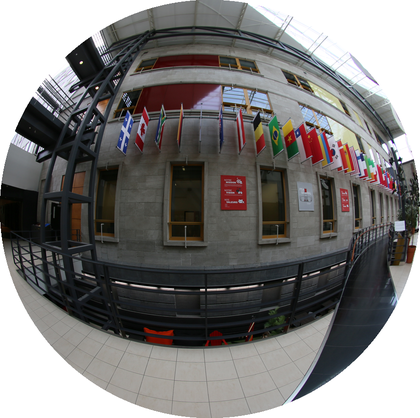}{0.3815}{0.0761}{0.6497}{0.3456} \\\addlinespace[2pt]
\qualrow{Ground truth} & \qualcrop{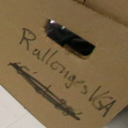}{128} & \qualcrop{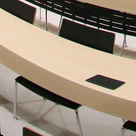}{136} & \qualcrop{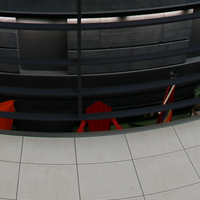}{200} \\\addlinespace[1pt]
\qualrow{Ours} & \qualcrop[676 FPS]{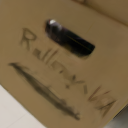}{128} & \qualcrop[642 FPS]{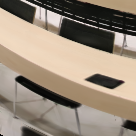}{136} & \qualcrop[578 FPS]{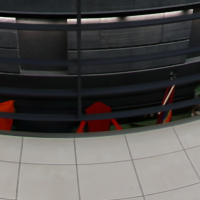}{200} \\\addlinespace[1pt]
\qualrow{DF-GS} & \qualcrop[488 FPS]{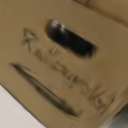}{128} & \qualcrop[401 FPS]{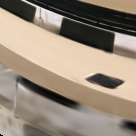}{136} & \qualcrop[361 FPS]{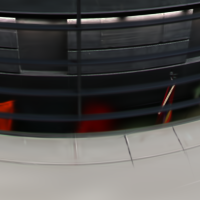}{200} \\\addlinespace[1pt]
\qualrow{3DGUT} & \qualcrop[276 FPS]{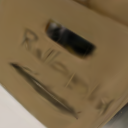}{128} & \qualcrop[266 FPS]{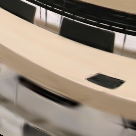}{136} & \qualcrop[254 FPS]{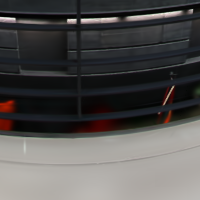}{200} \\\addlinespace[1pt]
\qualrow{3DGEER} & \qualcrop[144 FPS]{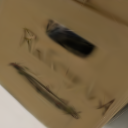}{128} & \qualcrop[127 FPS]{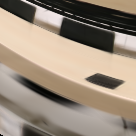}{136} & \qualcrop[123 FPS]{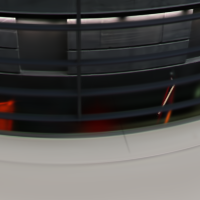}{200}
\end{tabular}}
\caption{Circular fisheye reconstruction on Fish\&Chips.}
\label{fig:native-fisheye-qualitative}
\end{minipage}\hfill
\begin{minipage}[t]{\columnwidth}
\centering
{\qualsetup
\begin{tabular}[t]{@{}c@{\hskip1pt}c@{\hskip1pt}c@{\hskip1pt}c@{}}
\qualstrut & \textsc{Lab} & \textsc{Flat1} & \textsc{Flat2} \\\addlinespace[2pt]
 & \qualview{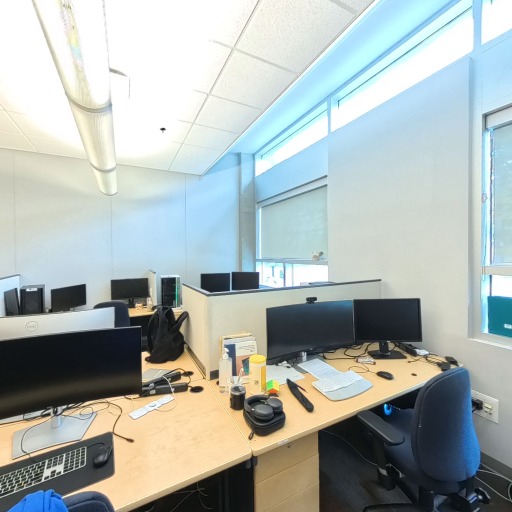}{0.3601}{0.0789}{0.6399}{0.3586} & \qualview{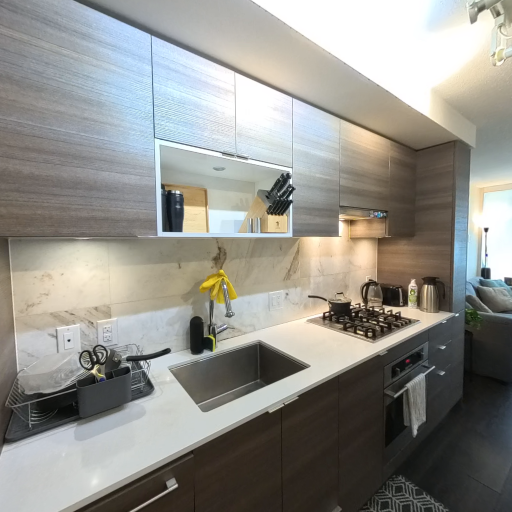}{0.3289}{0.1101}{0.6086}{0.3899} & \qualview{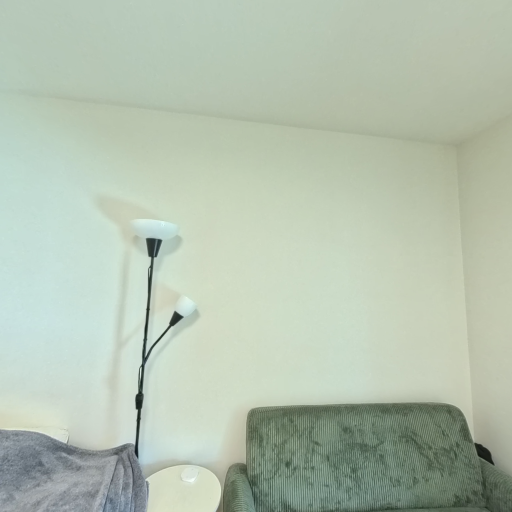}{0.2188}{0.1250}{0.5313}{0.4375} \\\addlinespace[2pt]
\qualrow{Ground truth} & \qualcrop{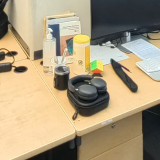}{160} & \qualcrop{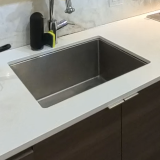}{160} & \qualcrop{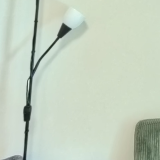}{160} \\\addlinespace[1pt]
\qualrow{Ours} & \qualcrop[1378 FPS]{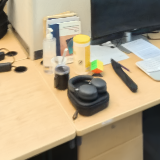}{160} & \qualcrop[1479 FPS]{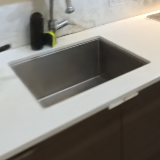}{160} & \qualcrop[1635 FPS]{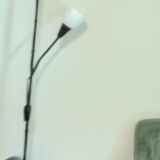}{160} \\\addlinespace[1pt]
\qualrow{FullCircle} & \qualcrop[293 FPS]{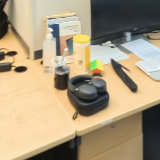}{160} & \qualcrop[324 FPS]{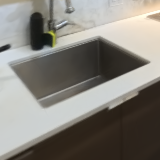}{160} & \qualcrop[266 FPS]{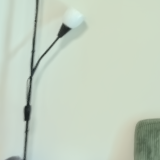}{160} \\\addlinespace[1pt]
\qualrow{SPaGS} & \qualcrop[851 FPS]{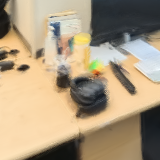}{160} & \qualcrop[894 FPS]{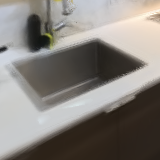}{160} & \qualcrop[801 FPS]{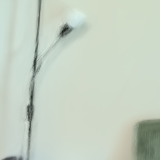}{160} \\\addlinespace[1pt]
\qualrow{UniTriSplat} & \qualcrop[28 FPS]{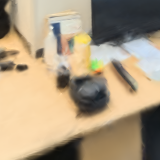}{160} & \qualcrop[35 FPS]{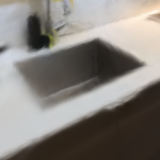}{160} & \qualcrop[32 FPS]{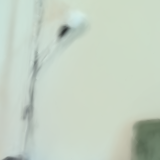}{160}
\end{tabular}}
\caption{Omnidirectional reconstruction on FullCircle~\cite{fullcircle}.}
\label{fig:native-omni-qualitative}
\end{minipage}
\end{figure*}

\paragraph{Results.} \Cref{tab:native-fisheye-results,fig:native-fisheye-qualitative} show that our method outperforms all rasterization baselines in PSNR, SSIM, LPIPS, rendering speed, and optimization time. While we found no significant benefit to using tangential or thin-prism distortion models, in scenes with close-up occluders, such as those in our Fish\&Chips dataset, a non-central lens model substantially improves reconstruction quality ($31.59 \rightarrow 32.37$\,dB PSNR, underlined in \cref{tab:native-fisheye-results}). Large improvements in specific image regions yield average PSNR gains of up to $+1.58$\,dB in some scenes (\cref{fig:fishchips-nearfield}), with notably more accurate reconstructions of objects near the camera (\cref{fig:noncentral-qualitative}).\footnote{Our calibration also recovers non-centrality in other datasets, but it has little effect on reconstructing scenes with only distant objects.}

\subsection{Stitchless omnidirectional reconstruction}
\label{sec:experiments:omni}

We evaluate omnidirectional reconstruction on the FullCircle~\cite{fullcircle} scenes, comparing against the FullCircle ray tracer~\cite{fullcircle} and the omnidirectional rasterizers SPaGS~\cite{spags} and UniTriSplat~\cite{unitrisplat}. FullCircle and our method train on raw dual-fisheye images, while SPaGS and UniTriSplat use stitched panoramas. We evaluate all methods on the same $512\!\times\!512$ virtual pinhole views with a 90\textdegree{} field of view, extracted at the center of held-out undistorted fisheye views.

\paragraph{Results.} As shown in \cref{fig:native-omni-qualitative}, training on the raw dual-fisheye images with ray tracing avoids the ghosting artifacts present in the stitched panoramas used by rasterization methods like SPaGS and UniTriSplat. Our method also renders nearly $6\times$ faster than FullCircle, while improving all metrics (\cref{tab:native-pano-results}).

\newpage
{\csname nolinenumbers\endcsname%
\noindent\begin{minipage}{\columnwidth}
\centering
{\footnotesize\setlength{\tabcolsep}{0pt}
\begin{tabular}{@{}c@{\hskip1pt}c@{\hskip1pt}c@{}}
Input (f/4) & Ours & Ground truth (f/11) \\
\includegraphics[width=\dimexpr(\columnwidth-2pt)/3\relax]{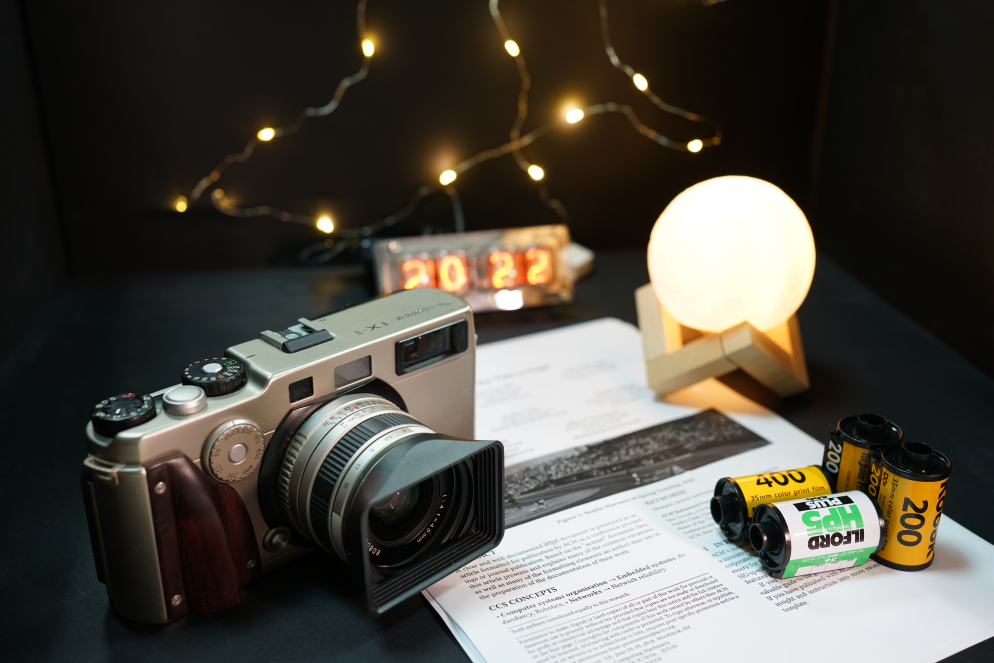} &
\includegraphics[width=\dimexpr(\columnwidth-2pt)/3\relax]{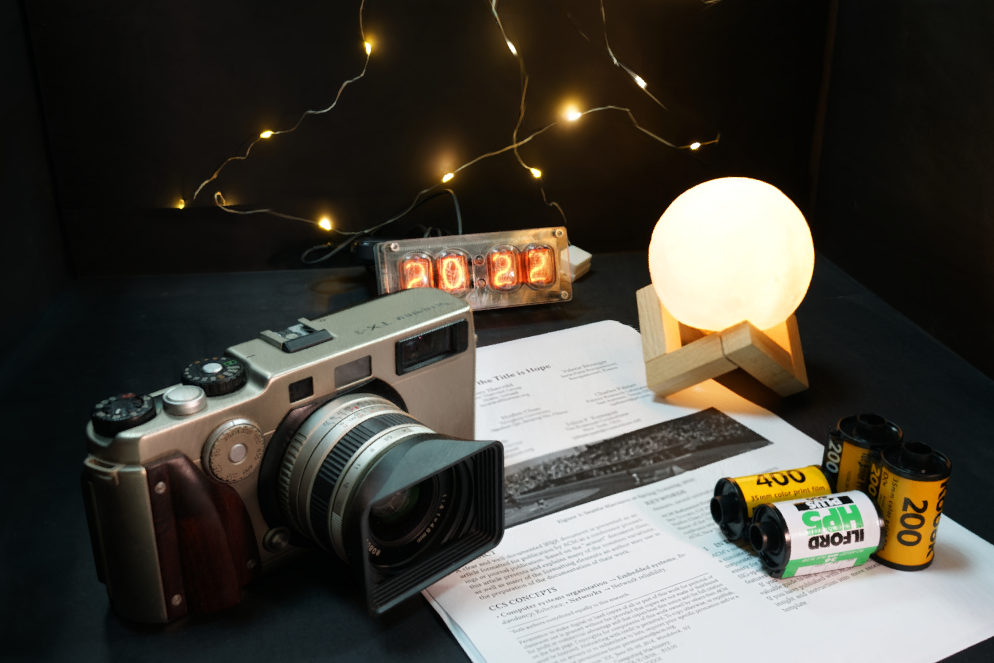} &
\includegraphics[width=\dimexpr(\columnwidth-2pt)/3\relax]{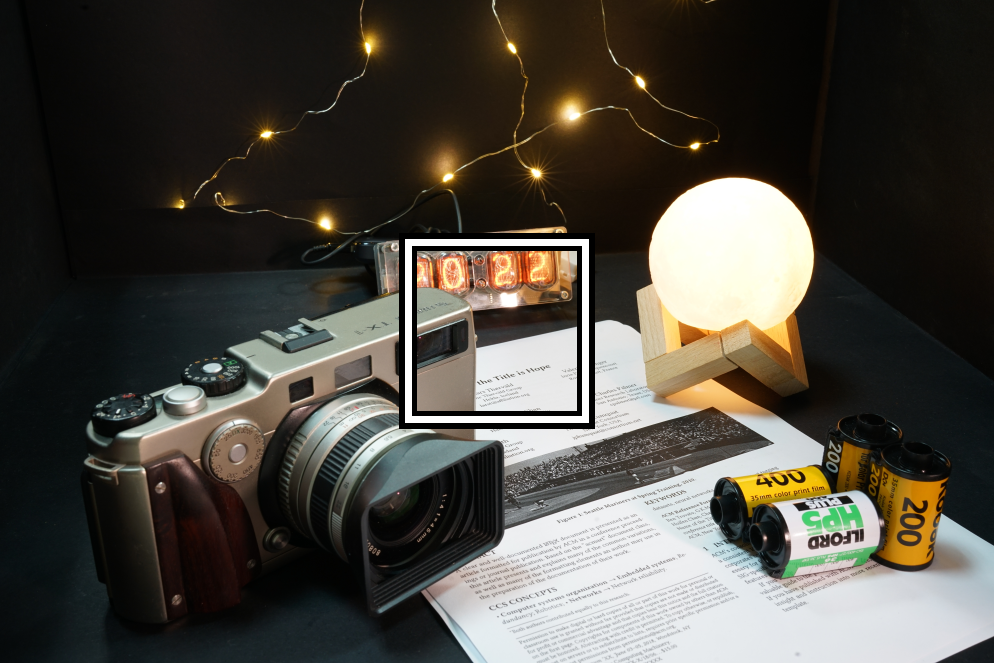} \\[-1pt]
\includegraphics[width=\dimexpr(\columnwidth-2pt)/3\relax,trim=0 17 0 0,clip]{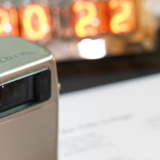} &
\includegraphics[width=\dimexpr(\columnwidth-2pt)/3\relax,trim=0 17 0 0,clip]{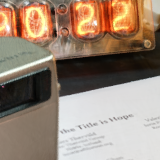} &
\includegraphics[width=\dimexpr(\columnwidth-2pt)/3\relax,trim=0 17 0 0,clip]{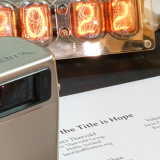}
\end{tabular}}
\captionof{figure}{Depth-of-field deblurring: training on f/4 inputs and recovering a sharp scene rendered at f/11 (supplemental).}
\label{fig:dof-qualitative}
\end{minipage}\par}
\newpage

\section{Discussion}
\label{sec:discussion}

Our work shows that Gaussian ray tracing can match or exceed specialized rasterizers in quality and speed across several wide-angle reconstruction scenarios; it can also model lens non-centrality, which is important in scenes with near-field occluders. While recent optimizations have substantially accelerated rasterization~\cite{higs,speedy-splat,faster-gs,fastgs} for pinhole cameras, we expect ray tracing's flexibility and conceptual simplicity to encourage wider adoption for wide-angle image reconstruction. The same ray-tracing framework naturally extends to depth of field (\cref{fig:dof-qualitative}), rolling shutter, motion blur, and complex camera systems, each of which would require a specialized rasterizer extension~\cite{dof-gs,gs-on-the-move,bad-gaussians,watersplatting}; please see the supplemental for additional experiments.

\clearpage

\section{LLM usage statement}
The authors used Claude 4.8--5.5 in the following ways:
\begin{itemize}
    \item Writing code and executing experiments.
    \item Literature search and mapping of related work.
    \item Generating Python plotting scripts used to create figures.
    \item Typesetting and LaTeX formatting.
    \item Drafting individual sentences on rare occasions.
    \item Brainstorming and proposing technical solutions.
    \item Cross-checking experiments, text, and references.
\end{itemize}
ChatGPT 5.6--6.1 was also used extensively for reformulating and improving writing.

{
    \small
    \bibliographystyle{ieeenat_fullname}
    \bibliography{references}
}

\end{document}